\documentclass[10pt, a4paper]{article}

\usepackage[final]{lrec2026} 

\usepackage{graphicx}   
\usepackage{caption}  
\usepackage{amsmath}

\usepackage{graphicx} 

\title{Improving low-resource ASR using bilingual fine-tuning with language identification: a cross-linguistic evaluation}

\name{Reihaneh Amooie$^{1}$, Yun Hao$^{1}$, Wietse de Vries$^{1}$, Jelske Dijkstra$^{2}$, Matt Coler$^{1}$ \\ {\bf \large 
Martijn Wieling$^{1,3}$}}

\address{$^{1}$University of Groningen, $^2$Fryske Akademy, $^3$Vrije Universiteit Brussel \\
         \{r.amooie, yun.hao, wietse.de.vries, m.coler, m.b.wieling\}@rug.nl, jdijkstra@fryske-akademy.nl}

\abstract{
This study explores how bilingual fine-tuning affects automatic speech recognition (ASR) in low-resource languages. We evaluate this method across nine linguistically and geographically diverse language pairs, covering a range of language families and writing systems. To distinguish the two languages, during training, we pre-pend  each input text with a language identification token. At inference, the model jointly predicts both the language and transcription from the speech input alone. As texts for which the language is incorrectly determined show low ASR performance, we also conduct a follow-up experiment in which the language identification token is provided both during training and inference. Our results show that bilingual fine-tuning can be beneficial when language identification accuracy is high, and that in cases where language identification performance is low, including the language identification token at inference helps to improve ASR performance. 
\\ \newline  }

\begin{document}

\maketitleabstract

\section{Introduction}
\label{sec:intro}

Automatic Speech Recognition (ASR) has progressed rapidly in recent years. However, much of that progress remains limited to high-resource languages, such as English. Low-resource languages still face challenges, including data scarcity and dialectal variation. A promising approach to address this challenge is cross-lingual transfer learning, where knowledge from high-resource languages supports low-resource ones. 


Both language-aware and language-agnostic multilingual fine-tuning strategies have been explored as promising approaches to improve ASR in low-resource settings by enabling cross-lingual transfer. \citet{yang2023learning}, for instance, proposed a sparse multilingual model with language-specific sub-networks and overlapping shared paths, allowing high-resource data to support lower-resource languages. Similarly, \citet{san2024predicting} demonstrated that supplementing a low-resource language with donor data can be highly effective: continued pre-training with only 10 hours of Punjabi plus 60 hours of Hindi nearly matched the performance of using 70 hours of Punjabi alone.

Another line of work explicitly incorporates language information. External language identification (LID) can aid disambiguation but increases latency \cite{waters2019leveraging}, leading to approaches that integrate implicit or explicit LID into ASR systems. Examples include meta-learning frameworks \cite{hsu2020meta,xiao2021adversarial}, adapter-based methods \cite{hou2020large,winata2020adapt}, and multi-task learning setups \cite{hou2020large,chen2015multitask}. In multi-task learning, \citet{chen2023improving} used an auxiliary CTC objective so that earlier encoder layers focused on language identification while later layers generated transcriptions conditioned on language identity. Relatedly, \citet{liu2023hierarchical} employed hierarchical Softmax with a Huffman Tree structure, exploiting similarities in linguistic unit frequency distributions across related languages to boost low-resource performance. A recent study by \citet{amooie2025evaluating} shows that Frisian ASR performance seems to improve when multilingual fine-tuning corpora are augmented with pre-pended language identification (LID) tokens, enabling the model to condition on language identity during fine-tuning and evaluation.

 However, it remains unclear whether bilingual fine-tuning with explicit language identification consistently improves ASR across diverse linguistic contexts. Prior work has not systematically examined this approach across language families, scripts, or varying degrees of linguistic similarity. 
 Factors such as linguistic proximity, writing script compatibility, and the model’s ability to disentangle languages have not been systematically examined. 
 In this work, we therefore investigate when and under what conditions bilingual fine-tuning using two related languages with explicit language identification (LID) tokens (prepended to each training sample to indicate its language) benefits low-resource languages.
 We evaluate the approach across nine typologically diverse pairs of related languages spanning five language families and multiple writing systems. For five target–donor pairs (for which the available data allows for sub-sampling), we train multiple models using random subsets of data to assess the robustness of any observed gains. 

\section{Data}

In addition to Frisian (with the aim of reproducing the results of \citealp{amooie2025evaluating}), we included eight additional target languages from five language families. 
We deliberately selected eight target languages representing diverse language families, typological features, and writing systems to evaluate the method across various low-resource scenarios.

For each target language, we first selected the most similar donor language available in Common Voice 17.0 ~\citet{ardila2020common} based on the lexical-phonetic distances (i.e.~the LDND distance) from the ASJP database~\cite[see also~\citealp{de2021adapting} for a similar approach]{wichmann2010evaluating}. 
We adopt a bilingual fine-tuning setup rather than a multilingual one, as prior work~\citealp{amooie2025evaluating} has shown that multilingual (involving more than two languages) fine-tuning does not clearly improve over bilingual fine-tuning.

All audio was extracted from Common Voice 17.0 and sampled at 16 kHz. To avoid bias toward higher-resource languages, we down-sampled both languages in each pair to 3,000 utterances. We control for the number of utterances (3,000 or fewer, depending on available resources) to ensure a comparable number of training instances across languages. While this does result in some variation of the total duration due to differences in utterance length distributions, we opted for consistency in the number of samples seen during training. When fewer utterances were available (i.e.~thereby representing a very-low-resource scenario), we matched both languages to the lower count. For five language pairs with over 3,000 utterances available (FY-NL, DA-SV, GL-IT, UK-BE, SK-CS), we repeated training 10 times using different random seeds to assess the statistical significance and robustness of the improvements. Table~\ref{tab:finetuning_datasets} shows all relevant information for each of these datasets.

Each model was evaluated on the test split of Common Voice 17.0 \citet{ardila2020common} for the target low-resource language in each pair. We did not down-sample the test sets. However, the development splits, which were used to monitor training progress and tune hyperparameters, were down-sampled to match the number of training samples.


\begin{table}[t]

\caption[skip=9pt]{Fine-tuning datasets (subsets from Common Voice 17.0; \citet{ardila2020common}).  Rows with a single language correspond to the target language, while rows with language pairs combine the target language (first) with a source language (second). Sim.~script: both languages use similar scripts; Dur.: duration; \# Utt.: number of utterances; \# Spk.: number of speakers.}
\label{tab:finetuning_datasets}
\centering
\scriptsize
\scriptsize
\scriptsize
\hspace*{-0.6cm} 
\renewcommand{\arraystretch}{1.5}
\setlength{\tabcolsep}{1.6pt} 
 \resizebox{\columnwidth}{!}{
\begin{tabular}{|l|l|c|c|c|c|}
\hline
\textbf{Languages} &
\textbf{ISO} & 
\textbf{Sim.~script} &
\textbf{Dur.} & 
\textbf{
\# Utt.} 
& \textbf{\# Spk.} \\
\hline
Frisian & FY &  - &  4.25h & 3000 & 192 \\
Frisian, Dutch & FY-NL &  yes & 8.27h & 6000 & 205 \\
\hline
Danish & DA & - & 3.44h & 3000 & 4   \\
Danish, Swedish & DA-SV & yes &  6.58h & 6000 & 20  \\

\hline
Galician & GL & - & 4.12h & 3000 & 26  \\
Galician, Italian &  GL-IT & yes & 8.21h & 6000 & 30  \\

\hline
Ukrainian & UK & - & 4.04h & 3000 &  29 \\
Ukrainian, Belarusian & UK-BE & yes & 8.15h & 6000 & 38  \\

\hline
Slovak &  SK & - &  3.21h & 3000 &  9  \\
Slovak, Czech & SK-CS & yes & 7.19h & 6000 &  20 \\

\hline
Serbian &  SR & -&  1.47h & 1880 &  6  \\
Serbian, Bulgarian & SR-BG & yes & 3.98h & 3760 & 8    \\

\hline
Slovenian &  SL & - & 1.34h & 1390 &  9  \\
Slovenian, Polish & SL-PL & yes & 3.19h & 2780 &  34  \\

\hline
Malayalam & ML & -& 1.42h & 1260 & 2  \\
Malayalam, Tamil & ML-TA & no & 3.28h & 2520 & 9  \\

\hline
Finnish & FI & -&  2.60h & 2080 & 6  \\
Finnish, Estonian & FI-ET & yes & 6.56h & 4160 & 240  \\



\hline
\end{tabular}
}
\end{table}

\subsection{Language pairs}

In this section, we briefly introduce the selected language pairs with linguistic information from Glottolog \cite{harald_hammarstrom_2024_14006617} and WALS \cite{matthew_dryer_2024_13950591}. 

\begin{itemize}
    \item \textbf{Germanic}
    
    \textit{Frisian} (target) and \textit{Dutch} (donor) were included as West Germanic languages to reproduce the results of \citet{amooie2025evaluating}. Both use the Latin alphabet (24 and 26 characters, respectively, depending on how they are counted), with Frisian featuring unique diacritics such as â, ê, and ô. Dutch was chosen as donor due to its high similarity to Frisian. 
    
    \textit{Danish} (target) and \textit{Swedish} (donor) represent North Germanic languages, both using the Latin alphabet (29 letters) but differing in their extra characters, e.g., ä and ö in Swedish versus æ and ø in Danish.


    \item \textbf{Romance}


    \textit{Galician} (target) and \textit{Italian} (donor) are Romance languages that both use the Latin script but follow different orthographic conventions. For instance, Galician includes the letter ñ and uses acute accents to mark non-default stress, whereas Italian lacks ñ but employs grave/acute accents (e.g., è, é, ò, ó), chiefly to indicate vowel quality and final stress.


    \item \textbf{Slavic}

    We included four pairs of Slavic languages that differ in script and spelling traditions. \textit{Ukrainian–Belarusian} and \textit{Serbian–Bulgarian} are written in Cyrillic, while \textit{Slovak–Czech} and \textit{Slovenian–Polish} use the Latin alphabet. Within each pair, the languages share the same script, but differ in orthographic conventions. For example, Ukrainian and Belarusian vary in certain characters and soft sign usage, while Serbian and Bulgarian differ in some Cyrillic letters and vowel notation.

\item \textbf{Dravidian}

\textit{Malayalam} (target) and \textit{Tamil} (donor) are Dravidian languages with writing systems that, while historically connected, developed distinct characteristics. Both scripts trace back to Brahmi, but Malayalam evolved through Grantha and retains more of its features, whereas Tamil underwent greater simplification and standardization. 

    \item \textbf{Uralic}

    \textit{Finnish} (target) and \textit{Estonian} (donor) are Uralic languages that use Latin alphabets with distinct diacritics. For example, Estonian employs š, ž, and o, which Finnish lacks, while Finnish uses å (mainly in Swedish loanwords). Estonian also omits c, q, w, x, and y.

\end{itemize}


\section{Method}

\subsection{Finetuning procedure}

We fine-tuned the pre-trained XLS-R 1B model~\cite{babu2021xls}, which is based on the Wav2Vec 2.0 architecture~\cite{baevski2020wav2vec} and contains a convolutional feature encoder followed by a transformer-based context network. 
During fine-tuning, we froze the convolutional feature encoder to retain the pre-trained acoustic representations and updated only the transformer layers. This approach aligns with standard practices in fine-tuning Wav2Vec 2.0 models for ASR tasks.
To train the models, we used a learning rate of 0.00008 and a batch size of 8 with 16 gradient accumulation steps. 
All experiments were conducted using 16-bit floating point precision on a single NVIDIA A100 GPU with 40 GB of RAM. 

We compared each model trained using the bilingual data to a baseline model trained only using the target language data. 
When the size of the dataset doubles, the training time necessarily increases. To ensure that any performance gain is caused by adding donor language data  rather than longer training, we held the number of training epochs constant (50) in all experiments. This was to ensure that every training example is seen 50 times in both monolingual and bilingual runs, so the bilingual model performs two times more updates only because it contains two times more training examples. This also prevents potentially overfitting the monolingual model.

\subsection{Language identification procedure}


To provide explicit language context during training and inference, we pre-pended a language identification (LID) token to each utterance (sentence) during training. This LID token serves as a ground-truth label indicating its language (e.g., [FY-NL] for Frisian, and [NL] for Dutch). 

In this approach, at inference, we do not provide the language identification token. The decoder first predicts the LID token, and then the transcription. By doing so, language identifications and transcriptions are learned and inferred jointly. 

\subsubsection{Follow-up experiment: Providing the correct LID during inference}

Proceeding from the assumption that samples for which the language was correctly predicted have better ASR performance, we also add a small experiment for a subset of language pairs in which we condition the model directly on the target language identity. To achieve this, we extend \texttt{Wav2Vec2ForCTC} with a simple language-specific bias embedding (one vector per language, sized to the vocabulary) and add this bias to the CTC logits at every time step. Each training and test utterance is assigned a numeric language ID (e.g., \texttt{0} for Danish, \texttt{1} for Swedish), which the model receives as an additional argument (\texttt{langid}) during  fine-tuning and inference. During both phases, the model looks up the bias vector corresponding to the given language. 
This bias vector functions as a language-specific prior that shifts the model's output distribution toward that language's characteristic phonemes and orthographic patterns. As a result, the decoder is more likely to predict characters and sequences typical of the target language.
The encoder remains completely shared across languages, so this mechanism introduces only a minimal number of additional parameters while explicitly informing the decoder about the correct language. 
This conditioning helps the decoder stay within the correct language space and prevents cross-language confusion during decoding.

\subsection{Evaluation metrics}

ASR performance is evaluated using the word error rate (WER). The benefit of bilingual fine-tuning is quantified as $\Delta$WER, defined as monolingual WER minus bilingual WER. Consequently, more positive values indicate a greater improvement (i.e.~reduced WER). LID performance was evaluated using accuracy, defined as the proportion of test utterances whose language was correctly identified as the target language.

\begin{table*}[t]
\captionsetup{font=footnotesize,skip=2pt}
\caption{WER comparison of the monolingual baseline vs.~the bilingual model. A positive $\Delta$WER indicates an improvement of the monolingual model compared to the bilingual model. For the five language pairs for which 10 random subsets of training data were used, the  $p$-value reflects the significance of a (two-tailed) single-sample $t$-test of the difference in WER with nine degrees of freedom. Significant $p$-values (< 0.05) are marked in boldface. Dist.~indicates the linguistic distance between the pairs, and LID acc.~indicates the accuracy of the inferred language identification across 10 random shuffles. 
}
\label{tab:wer_improvements}
\centering
\scriptsize
\renewcommand{\arraystretch}{1.5}
\setlength{\tabcolsep}{3pt}
\begin{tabular}{c|c|c|c|r|c|c}
\hline
\textbf{Language pair} & \textbf{WER}$_\textrm{monolingual}$ \textbf{(SD)} & \textbf{WER}$_\textrm{bilingual}$ \textbf{(SD)} & \textbf{$\Delta$WER (SD)} & \textbf{$p(t)$} & \textbf{Dist.} & \textbf{LID acc. (\%)} \\
\hline
FY--NL & 16.1 ($\pm$0.4) & 14.4 ($\pm$0.3)  &  +1.7 ($\pm$0.4) & \textbf{< 0.001} & 52.0 & 99.5 \\
DA--SV & 21.3 ($\pm$0.3) & 20.7 ($\pm$0.5)  &  +0.5 ($\pm$0.4)& \textbf{0.004}     & 52.4 & 96.5 \\
GL--IT & 10.8 ($\pm$0.1)   & 10.4 ($\pm$0.1)  & +0.4 ($\pm$0.2) & \textbf{< 0.001} & 49.9 & 99.5 \\
UK--BE & 27.9 ($\pm$0.9)   & 29.4 ($\pm$1.5) & -1.5 ($\pm$2.1) & 0.051                          & 48.1 & 90.9 \\
SK--CS & 25.2 ($\pm$0.4)  & 25.5 ($\pm$0.9)  & -0.3 ($\pm$1.0) & 0.319                         & 32.8 & 85.7 \\
\hline
SR--BG & 15.3 &  14.4 & +0.9  & N/A & 48.0  & 95.5 \\
SL--PL & 19.5  & 20.6 & -1.1  & N/A & 46.4  & 93.7 \\
FI--ET & 25.4  & 26.5 & -1.1 & N/A & 47.6 &  80.0 \\
ML--TA & 75.2  & 73.9 & +1.3 & N/A & 34.8 &  92.4\\
\hline

\end{tabular}%
\end{table*}

\section{Results and Discussion} 

\begin{figure}[h]
  \centering
  \includegraphics[width=\columnwidth]{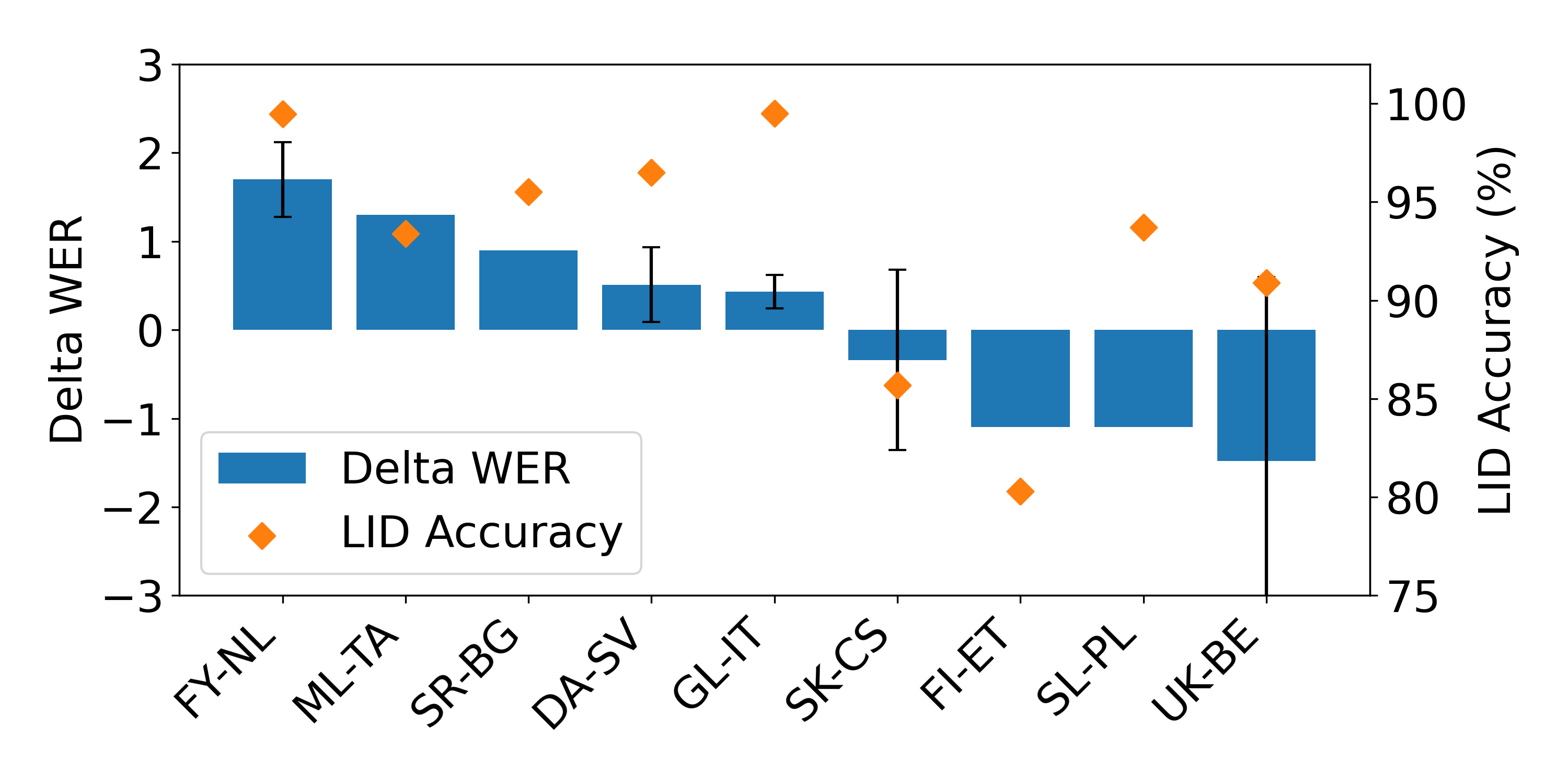}
    \vspace{-9mm}
   \caption{Average $\Delta$WER with error bars and LID Accuracy by language pair}
  \label{fig:lid_wer_overall}

\end{figure}

Table~\ref{tab:wer_improvements} and Figure~\ref{fig:lid_wer_overall} present the overall comparison of bilingual fine-tuning with language identification (LID) against monolingual baselines across all language pairs. For a subset of five language pairs (FY-NL, DA-SV, GL-IT, UK-BE, and SK-CS), we conducted multi-run experiments with 10 randomly selected subsets of 3000 training samples per language pair. As the remaining four language pairs (SR-BG, SL-PL, ML-TA, and FI-ET) had fewer than 3000 training samples per language pair available (i.e.~representing very-low-resource languages; see Table~\ref{tab:finetuning_datasets}) and subsetting was thus not possible, we only report the results of the single run including all data for these pairs. 

In Figure~\ref{fig:lid_wer_overall}, the blue bars denote the (average) improvement in word error rate ($\Delta$WER), while the orange markers indicate the corresponding LID accuracy. For the multi-run pairs, error bars show the variance across shuffles. For the remaining very-low-resource language pairs, results are based on a single run. The bars are sorted from left to right by decreasing $\Delta$WER. Overall, we observe a diverse pattern: a small majority of language pairs, such as Frisian–Dutch, Galician–Italian, and Danish–Swedish, show positive $\Delta$WER values, indicating consistent gains from bilingual fine-tuning, whereas a smaller number of pairs including Finnish–Estonian, Slovenian-Polish and Ukranian-Belarusian exhibit negative $\Delta$WER, suggesting that bilingual fine-tuning may also harm performance. However, it seems that bilingual fine-tuning only appears to harm performance when LID accuracy is not very high (i.e.~lower than 95\%). Furthermore, script differences do not appear to hinder transfer, as the ML-TA pair shows a positive $\Delta$WER despite differing scripts. Likewise, the number of training speakers also did not exhibit a strong effect, as Tamil has relatively few speakers in the training data, yet helped to improve ASR performance for Malayalam, whereas Estonian has many speakers in the training data, but did not improve ASR performance for Finnish. 

For the five language pairs for which multiple runs were possible (due to having more data to randomly sample from), Table~\ref{tab:wer_improvements} also summarizes the average differences in WER between the bilingual and monolingual models across the 10 shuffles. 
Single-sample $t$-tests of the differences in WER (H$_0$: $\Delta$WER$=0$) show that $\Delta$WER is significantly greater than zero for FY–NL, DA–SV, and GL–IT, but that there is no significant difference for the other two language pairs (UK-BE, and SK-CS). Consequently, for three out of five pairs, all showing an average LID accuracy over 95\%, the bilingual approach appears to be beneficial. This finding suggests that LID accuracy may be an important factor in determining ASR performance in our bilingual approach. In the following section, we will investigate this in more detail.  



\subsection{Relation between LID accuracy and $\Delta$WER}

\begin{figure}[tbp]
  \centering
  \includegraphics[width=\columnwidth]{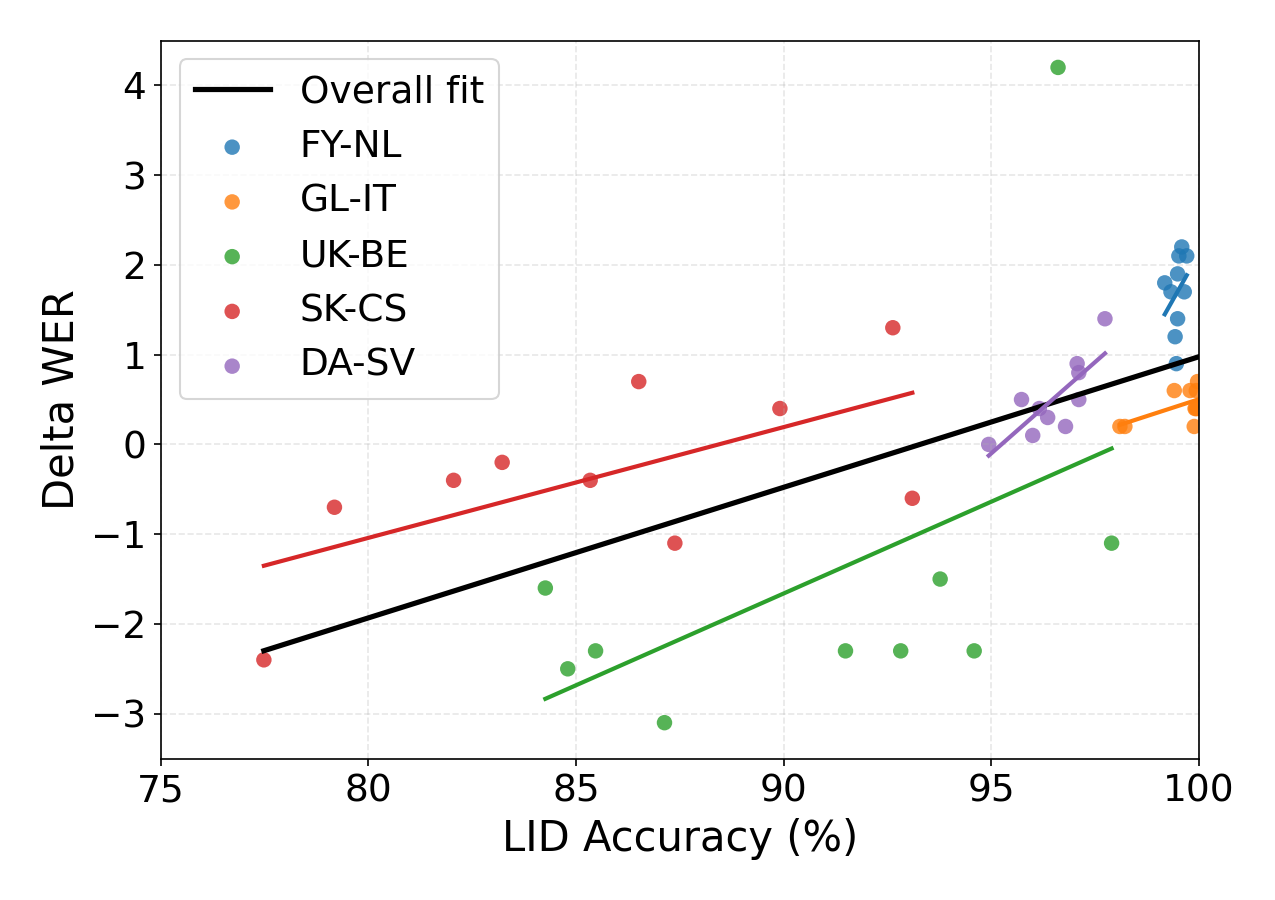}
  \vspace{-9mm}
  \caption{Relationship between LID accuracy (\%) and the performance difference between bilingual finetuning compared to the monolingual baseline.}
  \label{fig:lid_wer_scatter_with_lines}
  \vspace{-3mm}
\end{figure}

As language pairs with higher LID accuracy often exhibit a larger $\Delta$WER, we examine this relationship more systematically. Consequently, we determined the (Pearson) correlation between the LID accuracy and $\Delta$WER across the 10 random shuffles for the five language pairs included in Table~\ref{tab:wer_improvements}. Table~\ref{tab:correlation_results} reports the numerical results of this analysis, whereas Figure~\ref{fig:lid_wer_scatter_with_lines} provides the visualization in a scatter plot. The (regression) lines visualize the strength and direction of the correlation, both overall and per language pair. All correlations are positive (also reflected by the angle of the lines in Figure~\ref{fig:lid_wer_scatter_with_lines}). While the statistical significance varies by language pair, this is caused by the relatively low number of runs (i.e.~10) per language pair. The general pattern across all languages, however, is significant ($p$ < 0.001).\footnote{Note that due to the dependencies in our data (i.e.~multiple observations for five language pairs), the overall significance had to be determined via a linear mixed-effects regression analysis with language pair as a random-effect factor.\label{footnote_1}} 




\newcommand{\ninept}{\fontsize{7pt}{10.5pt}\selectfont}

\begin{table}[htbp]

\captionsetup{font=footnotesize,skip=2pt}

\caption{Pearson correlation analysis between $\Delta$WER and LID accuracy. Significant $p$-values (< 0.05) are marked in boldface.}
\label{tab:correlation_results}
\centering
\scriptsize
\setlength{\tabcolsep}{6pt}
\renewcommand{\arraystretch}{1.3}
\begin{tabular}{c|c|c|c}
\hline
\textbf{Language Pair} & \textbf{Pearson $r$} & \textbf{$p(r)$} & \textbf{$R^2$} \\
\hline
FY--NL & 0.298 & 0.403      & 0.089 \\
DA--SV & 0.792  & \textbf{0.006} & 0.627 \\
GL--IT & 0.582  & 0.078      & 0.338 \\
UK--BE & 0.501  & 0.141      & 0.251 \\
SK--CS & 0.642  & \textbf{0.045}  & 0.412 \\
ALL    & 0.621  & \textbf{< 0.001}\footref{footnote_1} & 0.386 \\
\hline
\end{tabular}

\end{table}

\begin{table}[htbp]

\captionsetup{font=footnotesize,skip=2pt}

\caption{WER for three language pairs under two proposed settings: Predicted LID at inference and Given LID at inference. For each setting we report the WER (\%). For the predicted LID, we provide two additional values between parentheses: the WER for the samples where the LID was correctly predicted (corr.), and the WER for the samples where the LID was incorrectly predicted (incorr.).}

\label{tab:corr_incorr_given_predicted}
\centering
\scriptsize
\setlength{\tabcolsep}{5pt}
\renewcommand{\arraystretch}{1.5}
\begin{tabular}{c|c|l}
\hline
\textbf{Pair} & \textbf{WER$_\textrm{Given LID}$} & \textbf{WER$_\textrm{Predicted LID}$ (corr.; incorr.)} \\
\hline
FY--NL & \textbf{14.3} & \textbf{14.3} (14.1; 53.3)  \\
DA--SV & \textbf{20.4}  & 22.2  (20.7; 59.7)   \\
SK--CS & \textbf{25.7}   & 26.3 (24.5; 35.4)  \\
\hline
\end{tabular}

\end{table}



\newcolumntype{C}[1]{>{\centering\arraybackslash}p{#1}}



These findings suggest that poor LID performance is a bottleneck limiting the benefits of bilingual fine-tuning. To test whether providing correct LID at inference could overcome this bottleneck, we conducted a follow-up experiment on three language pairs with varying LID accuracy.

\subsection{Providing LID at inference}

\label{sec:lid_filtering}
As higher LID generally appeared to be associated with a greater improvement in WER for the bilingual model, we also investigated a potentially beneficial effect of providing LID at inference. For this, we conducted an additional experiment for three language pairs (FY-NL, DA-SV, and SK-CS) for which 3000 samples per language were available. For each language pair, we conducted a single experiment to limit the required computational time. 

The potential benefit of providing LID during inference will both be dependent on the LID performance for a language pair, as well as the difference in WER for samples for which the LID was identified correctly versus those for which the LID was identified incorrectly. Specifically, for language pairs with a lower LID performance, and those with a greater difference in WER between correctly versus incorrectly identified samples, there will be a greater potential for improvement. 

The three language pairs we selected varied along these dimensions. Specifically, LID performance for FY-NL was very high (over 99\%), whereas it was lower for DA-SV (about 95\%), and lowest for SK-CS (about 89\%). The WER of samples for which the LID was incorrectly identified was always higher than for those whose LID was correctly identified, but this performance gap also differed per language pair. For FY-NL and DA-SV, the difference was large (about 40\%; 14.1\% for FY-NL samples with a correct LID vs.~53.3\% for FY-NL samples with an incorrect LID, and 20.7\% vs.~59.7\%, respectively, for DA-SV), whereas it was much smaller for SK-CS (about 11\%; 24.5\% vs.~35.4\%, respectively), likely due to the increased similarity between the two languages (see Table~\ref{tab:wer_improvements}). 

Taking these two aspects together, we expect a greater potential for improvement for both DA-SV and SK-CS compared to FY-NL. Indeed, our results shown in Table \ref{tab:corr_incorr_given_predicted} show this assumption to be correct. Due to the very high LID, explicitly including LID at inference for FY-NL resulted in equal performance (14.3\%) compared to not including it. By contrast, for both DA-SV (WER reduction of 1.8\%; from 22.2\% to 20.4\%), as well as SK-CS (WER reduction of 0.6\%; from 26.3\% to 25.7 \%), the results improved by explicitly including LID at inference.

\subsection{Limitations}
While we identified clear correlation between ASR and LID performance, we did not investigate why LID performance was lower for some language pairs than for others. While we expect this pattern to be similar across model architectures, we did not investigate the effect of model architecture, the fine-tuning approach we used, or the stopping criterion. Likewise, we did not investigate whether the bilingual approach, in cases where the LID was correctly identified, may have resulted in new mistakes compared to the monolingual model. We will address these questions in future work. It is clear, however, that a lower LID performance will result in performance degradation, as the WER will be much higher for samples where language was incorrectly identified (as is shown in Table~\ref{tab:corr_incorr_given_predicted}).  While adding a third or even a fourth language has shown to improve results slightly \cite{amooie2025evaluating}, it also tends to result in lower LID performance. Manually providing LID for each sample solves this issue, but may not always be possible in automated pipelines.

\section{Conclusion}

In this study, we investigated bilingual fine-tuning with LID tokens for low-resource ASR across nine diverse language pairs. Our analysis shows that performance gains are not consistent across all settings but are strongly associated with LID accuracy. Reliable language disambiguation, particularly in related languages, leads to clear improvements, while lower LID accuracy limits or even counteracts the benefits, as the ASR performance is lower for samples with an incorrect LID. In that case, preliminary results showed that providing the correct language identification at inference helps to improve performance compared to the monolingual baseline. Consequently, in cases of little data for one language, ASR performance for that language may be increased by supplementing the monolingual data with data of a similar language, as long as LID tokens are provided, or LID identification performance is sufficiently high (i.e.~over 95\%). In sum, through this study, we have extended prior evidence from Frisian \cite{amooie2025evaluating} to a broader set of languages and show the benefit of bilingual training with language identification for low-resource ASR. 




\section{Bibliographical References}\label{sec:reference}

\bibliographystyle{lrec2026-natbib}
\bibliography{lrec2026-example}


\end{document}